\documentclass{article}

 \usepackage[main, final]{tagds_2025}

 \workshoptitle{Topology, Algebra, and Geometry in Data Science}

\usepackage[utf8]{inputenc} 
\usepackage[T1]{fontenc}    
\usepackage{hyperref}       
\usepackage{url}            
\usepackage{booktabs}       
\usepackage{amsfonts}       
\usepackage{nicefrac}       
\usepackage{microtype}      
\usepackage{xcolor}         
\usepackage{subcaption}
\usepackage{amsmath}
\usepackage{enumitem}
\usepackage{comment}
\usepackage[dvipsnames]{xcolor} 

\usepackage[most]{tcolorbox} 

\newcommand{\Zp}{\mathbb{Z}/p\mathbb{Z}}

\newtcolorbox{mybluebox}{
    enhanced,              
    boxrule=0.5pt,         
    rounded corners,       
    arc=4mm,               
    colback=blue!10!white, 
    colframe=blue!75!black 
}

\title{Can Neural Networks Learn Small Algebraic Worlds? An Investigation Into the Group-theoretic Structures Learned By Narrow Models Trained To Predict Group Operations}

\author{%
   Henry Kvinge$^{1,2}$, Andrew Aguilar$^{1,2}$, Nayda Farnsworth$^{1,3}$, Grace O'Brien$^{1,2}$, \\\textbf{Robert Jasper}$^{1}$, \textbf{Sarah Scullen}$^{1}$, \textbf{Helen Jenne}$^{1}$\\
  $^1$Pacific Northwest National Laboratory\\
  $^2$Department of Mathematics, University of Washington\\
  $^3$Colgate University\\
  \texttt{henry.kvinge@pnnl.gov}\\
}

\begin{document}

\maketitle

\begin{abstract}
While a real-world research program in mathematics may be guided by a motivating question, the process of mathematical discovery is typically open-ended. Ideally, exploration needed to answer the original question will reveal new structures, patterns, and insights that are valuable in their own right. This contrasts with the exam-style paradigm in which the machine learning community typically applies AI to math. To maximize progress in mathematics using AI, we will need to go beyond simple question answering. With this in mind, we explore the extent to which narrow models trained to solve a fixed mathematical task learn broader mathematical structure that can be extracted by a researcher or other AI system. As a basic test case for this, we use the task of training a neural network to predict a group operation (for example, performing modular arithmetic or composition of permutations). We describe a suite of tests designed to assess whether the model captures significant group-theoretic notions such as the identity element, commutativity, or subgroups. Through extensive experimentation we find evidence that models learn representations capable of capturing abstract algebraic properties. For example, we find hints that models capture the commutativity of modular arithmetic. We are also able to train linear classifiers that reliably distinguish between elements of certain subgroups (even though no labels for these subgroups are included in the data). On the other hand, we are unable to extract notions such as the concept of the identity element. Together, our results suggest that in some cases the representations of even small neural networks can be used to distill interesting abstract structure from new mathematical objects.
\end{abstract}

\section{Introduction}

Deep learning-based systems are increasingly being used as a tool to accelerate research mathematics. Though there is a growing body of work that aims for generalist AI scientists \cite{lu2024ai, yamada2025ai} or program synthesis systems like AlphaEvolve \cite{novikov2025alphaevolve,romera2024mathematical}, the majority of AI for math work still starts with a specific problem of interest and then builds a system to learn a solution to this problem. This system may be a narrow model trained exclusively on a task related to the problem \cite{Davies2021AdvancingMB} or it may be a more sophisticated framework using foundation models like LLMs. What these set-ups have in common is that they are often restricted to revealing solutions to the initial question. Real mathematics research on the other hand is substantially more open-ended with the final output of a research program often varying substantially from the initial motivating question.  

If the goal is to develop methodologies that enable more open-ended discovery, one potential solution is to look more closely at effective narrow models. Might these already contain valuable insights that were learned in the course of solving the original task? There are a range of case studies that provide precedent for such a hypothesis. For instance, careful analysis of models trained to perform modular arithmetic \cite{nanda2023progressmeasuresgrokkingmechanistic} or composition of permutations \cite{chughtai2023toy,  stander2023grokking, wu2024unifying} have revealed sophisticated algorithms that depend on the representation theory of the corresponding groups. Similarly, analysis of a graph neural network designed to perform classification of the mutation equivalence class of a finite or affine type quiver was found to naturally cluster instances in ways that align with human classification schemes \cite{hemachines}.

Motivated by this question, we explore the elementary case of a neural network trained to predict the operation of a group $G$ and ask to what extent we can detect basic group-theoretic concepts from such a network. Notions we consider include commutativity, the identity, and subgroups, all core concepts within group theory. We explore three approaches to detecting these concepts: (i) through training dynamics where we look for changes in loss/accuracy that might correspond to a model learning a new concept, (ii) differences in performance across subsets of input instances (for example, a model that `understands’ the identity element $e$ should always get the correct answer on questions of the form $g \star e$ or $e \star g$, even when it has never seen $g$ before), and (iii) the structure of the internal representations of the group. 

We probe MLPs and transformers trained to perform the group operation for cyclic groups, symmetric groups, and dihedral groups of varying sizes. Across these settings we offer evidence that at least some of the mathematical concepts that we study are captured by these simple models. For example, for some subgroups $H$ linear probes are able to reliably distinguish between pairs of elements $(g_1,g_2)$ which both belong to $H$ and pairs where either $g_1$ or $g_2$ does not belong to $H$. This is significant since these models were trained without any information about subgroups. On the other hand, we failed to detect other concepts such as the significance of the identity element. Whether this is because this concept was never learned by the model or it is too nuanced for us to easily capture using our existing set of model analysis methods remains unclear.

All of this suggests that the {\emph{mathematical world models}} learned by small neural networks (easily accessible to even those with very restricted compute budgets) can contain interesting mathematical insights if one is willing to put in the work to extract them.
In summary our contributions include the following.
\begin{itemize}[leftmargin=.5cm]
\item We describe a framework that uses finite groups and basic notions from group theory to better understand whether narrowly trained neural networks have world models sufficiently rich for open ended mathematical discovery.
\item We evaluate a range of approaches to extract the fingerprints of concepts such as commutativity or the notion of a subgroup from a neural network trained for a fixed task.
\item We provide conjectures about the type of concepts most likely to be captured by narrow models.
\end{itemize}

\section{Finite Groups and Their Associated Concepts}

The notion of a group is a central concept in modern mathematics. A group is a set $G$ along with a binary operation $\star: G \times G \rightarrow G$ which satisfies the following axioms. (1) $\star$ is associative so that for all $g_1,g_2,g_3 \in G$, $(g_1 \star g_2) \star g_3 = g_1 \star (g_2 \star g_3)$. (2) There is an identity element $e \in G$ such that $g \star e = e \star g = g$ for all $g \in G$. (3) Each $g \in G$ has an inverse element $g^{-1} \in G$ such that $g \star g^{-1} = e = g^{-1} \star g$. 

Despite arising from only three simple axioms, groups exhibit amazingly rich structure ranging from cyclic groups (familiar from modular arithmetic) to the {\emph{monster}}, the largest sporadic group which has order $\approx 10^{53}$. Careers have been spent studying groups and one product of this is a rich language that captures the breadth of structure arising in this field. In this work we only scratch the surface of this, asking whether we can recover the following notions from a neural network trained to perform the binary operation $\star$:
\begin{itemize}[leftmargin=.5cm]
    \item \textbf{Commutativity of the binary operation $\star$:} $\star$ is commutative if for all $g_1, g_2 \in G$, we have $g_1 \star g_2 = g_2 \star g_1$.
    \item \textbf{The identity element:} The element $e$ is uniquely defined in $G$ by the fact that $e \star g = g = g \star e$ for all $g \in G$.
    \item \textbf{Subgroup structure:} There may exist proper, non-trivial subsets $H \subseteq G$ that are closed under $\star$ and hence form groups in their own right. These {\emph{subgroups}} form lattices under the containment relationship.
\end{itemize}

We look at three different families of groups, which we describe here.

\noindent\textbf{Cyclic groups, $\Zp$:} Cyclic groups are familiar since they correspond to modular addition. We can represent $\Zp$ with the elements $\{\overline{0},\overline{1}, \dots, \overline{p-1}\}$ and realize the binary operation as $\overline{a} + \overline{b} = \overline{a + b \mod p}$. $\Zp$ is commutative and has order $|\Zp| = p$. The subgroups of $\Zp$ are in one-to-one correspondence with integers $1 \leq k \leq p$ such that $k$ divides $p$. If $k$ divides $p$, then we can realize the corresponding subgroup as $\{\overline{0}, \overline{k}, \overline{2k}, \dots \}$.

\noindent\textbf{Symmetric groups, $S_n$:} The symmetric group $S_n$ is the set of all permutations of $n$ elements with the binary operation of composition of permutations. As such, the order of $S_n$ is $n!$. It is not commutative. $S_n$ has many subgroups but we will work with one of the most well-known, the {\emph{alternating (sub)group}}, which consists of all permutations of $n$ which are even. The alternating group has size $n!/2$.

\noindent\textbf{Dihedral groups $D_n$:} The dihedral group $D_n$ can be realized as the set of rotations and reflections that preserve the $n$-gon. It consists of $n$ rotations and $n$ reflections, making it a group of order $2n$. Subgroups of $D_n$ include the subgroup of all $n$ rotations.

\section{How Can We Detect Whether a Model Has Learned a Mathematical Concept?}
\label{sec:detection-approaches}

Suppose $f: G \times G \rightarrow G$ is a neural network that has been trained to perform the binary operation $\star$ of a group. Thus, provided with $g_1,g_2 \in G$, $f$ predicts $g_1 \star g_2$. We outline the three broad approaches that we use to detect whether $f$ has `learned' algebraic concepts that characterize groups (at least, as a human mathematician would describe them). 
\begin{itemize}[leftmargin=*]
    \item {\textbf{Learning dynamics:}} Detailed analysis of neural network training has revealed that (at least in simple problems), sudden drops in loss may correspond to a network gaining a specific capability \cite{chensudden,olsson2022context}. Might we see similar changes in the loss curve when a network trained on a group binary operation learns a concept like commutativity of $\star$? Motivated by this idea, we explore whether changes in accuracy or loss correlate with changes in model performance on specific subpopulations of the test set which capture a certain concept. For example, one can imagine that a sudden drop in loss might correspond to a model achieving high accuracy on instances of the form $e \star g$ or $g \star e$, suggesting that the model has learned the concept of the identity element. 

    \item {\textbf{Generalization:}} Mathematical concepts are valuable precisely because they allow us to reason broadly across instances we have not seen before. One may not have ever actually worked with the numbers $2,483,402$ and $5,840,202$ but we can immediately say that $2,483,402 + 5,840,202 = 5,840,202 + 2,483,402$ based on mathematical concepts that we already understand. One way to evaluate whether a model has learned a concept is to see whether the model can apply that concept to an out-of-distribution example.
    
    \item {\textbf{Structure of Internal Representations:}} A model’s understanding of a concept may manifest as structure in the hidden activations of the model. For example, a model may encode commutativity by representing $g_1 \star g_2$ and $g_2 \star g_1$ as more similar than arbitrary $g_1 \star g_2$ and $g_3 \star g_4$ even when $g_1 \star g_2 = g_3 \star g_4$. This perspective aligns with the mechanistic interpretability paradigm.
\end{itemize}

\subsection{Experimental details}
\label{sec:details}

In our experiments we use MLPs and transformers that are of a scale that is accessible to most researchers but are sufficient to learn the group operation. Our MLPs consist of dense linear layers interleaved with ReLU nonlinearities. Our transformers are decoder-only and use GeLU nonlinearities. The task is framed as prediction and thus uses a standard cross-entropy loss function. All models are trained using the Adam optimizer with varying learning rate  values and weight decay on a single Nvidia A100. In most experiments, we explore a wide range of hyperparameters. We provide the hyperparameters that we use for the paper's visualizations in Section~\ref{sect:hyperparameters} in the Appendix.

Our experiments use a one-hot encoding of elements of $G$. For MLPs, we encode $g_1 \star g_2$ by concatenating two length $|G|$ vectors into a single $2|G|$-dimensional input. Since the output prediction is an element of $G$, the output dimension is $|G|$. For transformers we encode $g_1 \star g_2$ as a length 3 sequence, the first token corresponding to $g_1$ and the second token corresponding to $g_2$. The final token, on which the transformer's prediction is made, can be taken to correspond to `$=$'. Thus, the transformer operates on a vocabulary of size $|G|+1$. In our experiments, we work with the cyclic groups $C_{64}$, $C_{67}$, $C_{100}$, $C_{256}$, $C_{257}$, $C_{508}$, $C_{512}$, the symmetric groups $S_4$, $S_5$, and $S_6$, and the dihedral groups $D_{30}$, $D_{50}$, $D_{60}$, $D_{120}$, and $D_{240}$.

\subsection{Commutativity}
\label{sect-comm}

It is easy to see why knowing that a group is commutative can lead to more efficient computation. If $G$ is commutative, one immediately knows the value of $g_2 \star g_1$ once they know the value of $g_1 \star g_2$. Beyond this, commutativity has deep consequences for the types of structural features that a group can exhibit with commutative groups generally being much simpler. Our first set of experiments aim to understand whether MLPs and transformers capture commutativity using the perspectives described at the beginning of Section~\ref{sec:detection-approaches}. Our work extends \cite{kvinge2025probing}, which used the cosine similarity test below to determine whether large language models have an internal notion of commutativity. We begin by describing our experiments.

\textbf{Symmetric consistency:} As noted above, one sign that a model has internalized the notion of commutativity would be that the model's prediction of $g_1  \star g_2$ and $g_2 \star g_1$ will tend to be the same, regardless of correctness. The following quantities aim to measure this. 

Let $\mathcal{S}$ be the full set of pairs $\big((g_1,g_2),(g_2,g_1)\big)$ in the test set. The {\emph{symmetric consistency}} is measured by computing the fraction of pairs where $f(g_1,g_2) = f(g_2,g_1)$. 
\begin{equation*}
\text{Consistency of $f$} := \frac{1}{|\mathcal{S}|}\#\{f(g_1,g_2) = f(g_2,g_1) \; | \; (
(g_1,g_2),(g_2,g_1)) \in 
\mathcal{S}\}.
\end{equation*}
Symmetric consistency values closer to 1 indicate that $f$ makes more consistent predictions across pairs $(g_1,g_2)$ and $(g_2,g_1)$. However, care is needed because this consistency statistic alone can be maximized through better performance on the test set. In other words, any model that achieves 100\% accuracy on the test examples will have symmetric consistency equal to 1, even if it is simply a look-up table. To mitigate this, we also measure {\emph{equal value consistency}} which is the fraction of times that $f$ predicts $f(g_1,g_2) = f(g_3,g_4)$ for $(g_1, g_2), (g_3,g_4) \in S$ such that $g_1 \star g_2 = g_3 \star g_4$. We compare these over the course of training. Increases in symmetric consistency independent of equal value consistency may be evidence of a learned concept of commutativity in the model.

Another way we can probe for commutativity is to look at how strongly symmetric consistency holds on out-of-distribution examples that were not seen during training. We call this version {\emph{out-of-distribution (OOD) symmetric consistency}}. Here we choose some $g' \in G$ and remove all examples from $S_{g'} := \{(g',g)$,$(g,g') \;|\; g \in G\}$ from the training set. We then measure symmetric consistency on $S_{g'}$ to see if the model can apply any learned notion of commutativity to elements of $S_{g'}$ which feature the unseen element $g'$. 

\textbf{Cosine similarity:} Commutativity means that, algebraically, we are allowed to treat $g_1 \star g_2$ and $g_2 \star g_1$ as equal. In the context of large language models, \cite{kvinge2025probing} hypothesized that such an understanding might manifest as the model constructing internal representations of $g_1 \star g_2$ and $g_2 \star g_1$ that are closer in hidden activation space. For a given input $(g_1,g_2)$, denote by $v^k_{g_1,g_2}$ an activation vector corresponding to $(g_1,g_2)$ extracted from the $k$th layer of $f$. The {\emph{symmetric representational similarity of $f$ at layer $k$}} is
\begin{equation}\label{eqn-symmetric-representational-similarity}
\text{Symmetric representational similiarity of $f$} := \frac{1}{|S|}\sum_{(g_1,g_2),(g_2,g_1) \in \mathcal{S}} \text{Sim}(v_{g_1,g_2}^k,v_{g_2,g_1}^k).
\end{equation}
As in the case of symmetric consistency, we also compute a version of \eqref{eqn-symmetric-representational-similarity} where the sum is taken over pairs $(g_1,g_2), (g_3,g_4)$ where $g_1 \star g_2 = g_3 \star g_4$ to ensure trends that we see do not simply arise because $g_1 \star g_2 = g_2 \star g_1$.

\textbf{Do models capture commutativity?:} The only class of commutative groups that we explored were cyclic groups, so we focus our analysis on this setting. We note that by design, transformers are better adapted for capturing commutativity since the representations of the tokens corresponding to $g_1$ and $g_2$ in $g_1 \star g_2$ are the same as the representations of the tokens corresponding to $g_1$ and $g_2$ in $g_2 \star g_1$ up to modification by positional encoding. On the other hand, the one-hot encodings of $g_1 \star g_2$ and $g_2 \star g_1$ are generally orthogonal. 

For the first few epochs after initialization models have high symmetric consistency, equal value consistency, and OOD symmetric consistency since they tend to predict the same (incorrect) class for all input pairs. This phase passes as all types of consistency drop, with equal value consistency dropping much lower than symmetric consistency. In the non-OOD setting, we then see both symmetric and equal value consistency increase again until they are both near 1 (corresponding to 100\% test accuracy). Besides dropping less than equal value consistency, symmetric consistency also increases more quickly than equal value consistency across runs. This suggests that models may capture $f(g_1,g_2) = f(g_2,g_1)$ via a distinct mechanism relative to $f(g_1,g_2) = f(g_3,g_4)$ when $g_1 \star g_2 = g_3 \star g_4$. Such a distinct mechanism can itself be interpreted as evidence of a concept of commutativity. 

\begin{figure}
    \centering
        \includegraphics[width = .49\textwidth]{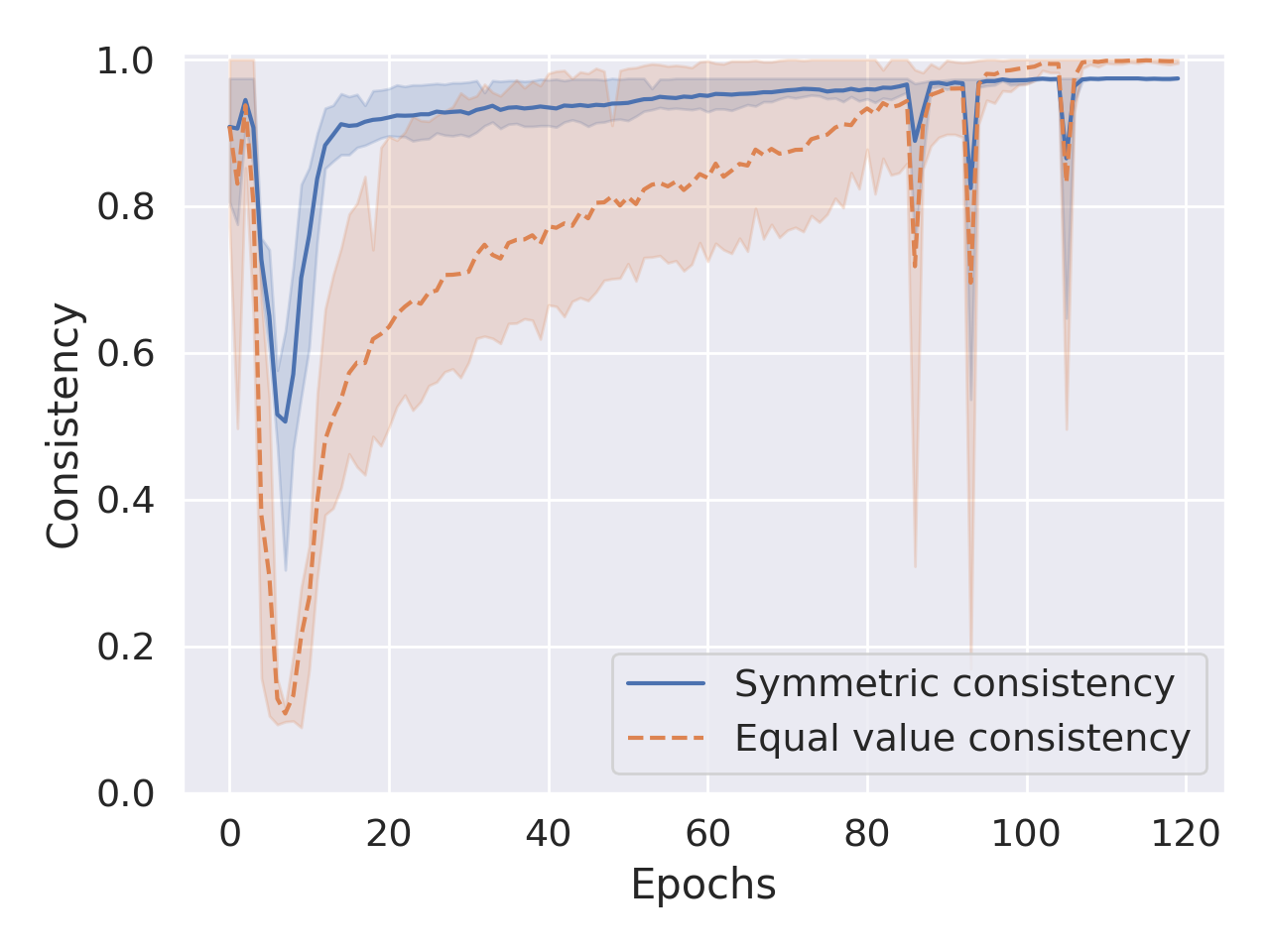}
        \includegraphics[width = .49\textwidth]{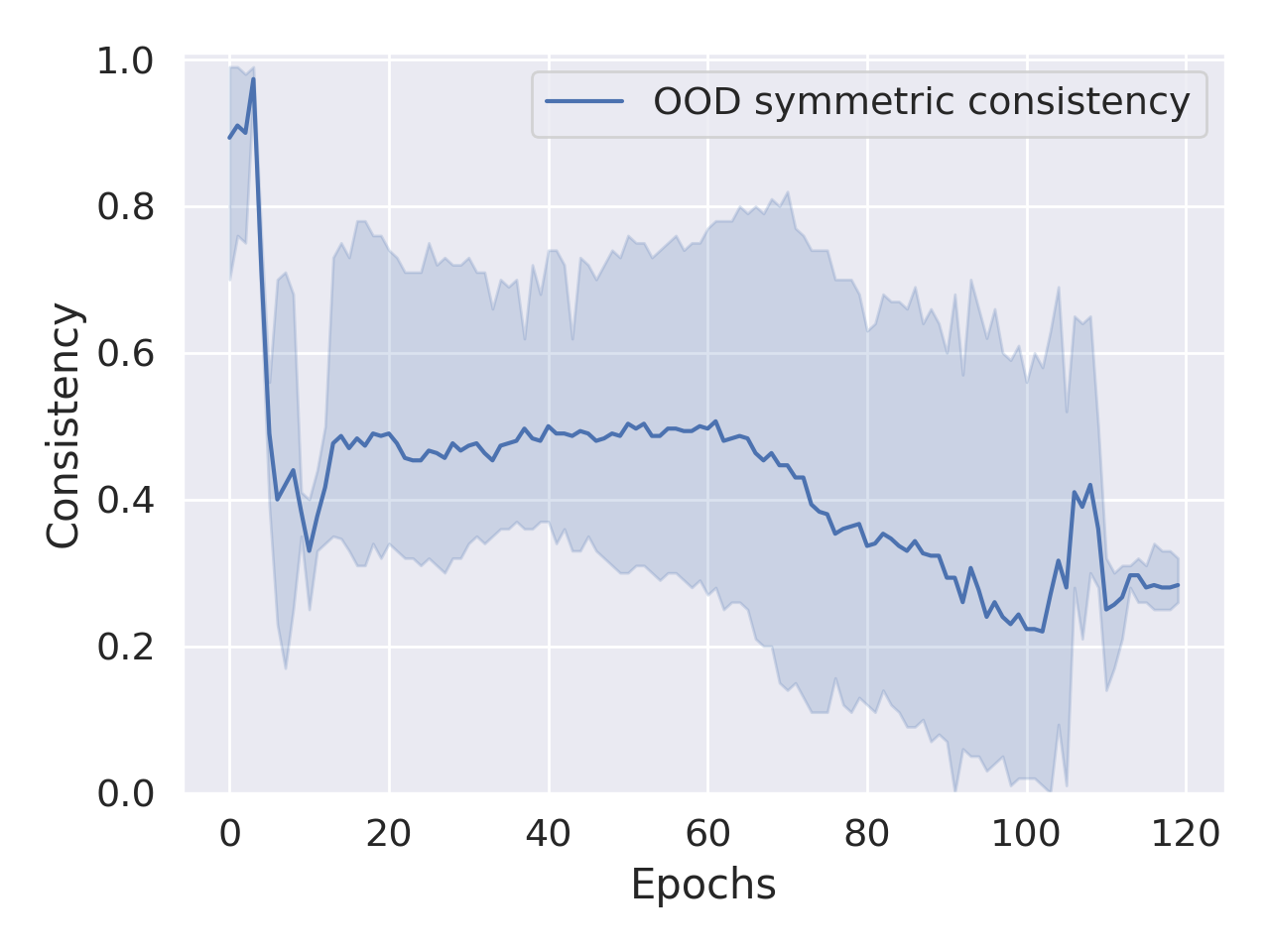}
    \caption{Plots of symmetric consistency vs. equal value consistency \textbf{(Left)} and OOD symmetric consistency vs. \textbf{(Right)} for five transformers trained to perform modular arithmetic in $C_{100}$. Shaded regions correspond to $95\%$ confidence intervals.}
    \label{fig:commutativity}
\end{figure}

In the OOD setting, symmetric consistency values do not increase again after their initial drop, but remain elevated above what we would expect if these models were simply guessing the value of $g' \star g$ and $g \star g'$ independently, which is $1/|G|$. This is visualized in Figure \ref{fig:commutativity} for five transformers trained to predict the binary operation of the cyclic group $C_{100}$. 

\begin{figure}
    \centering
        \includegraphics[width = \textwidth]{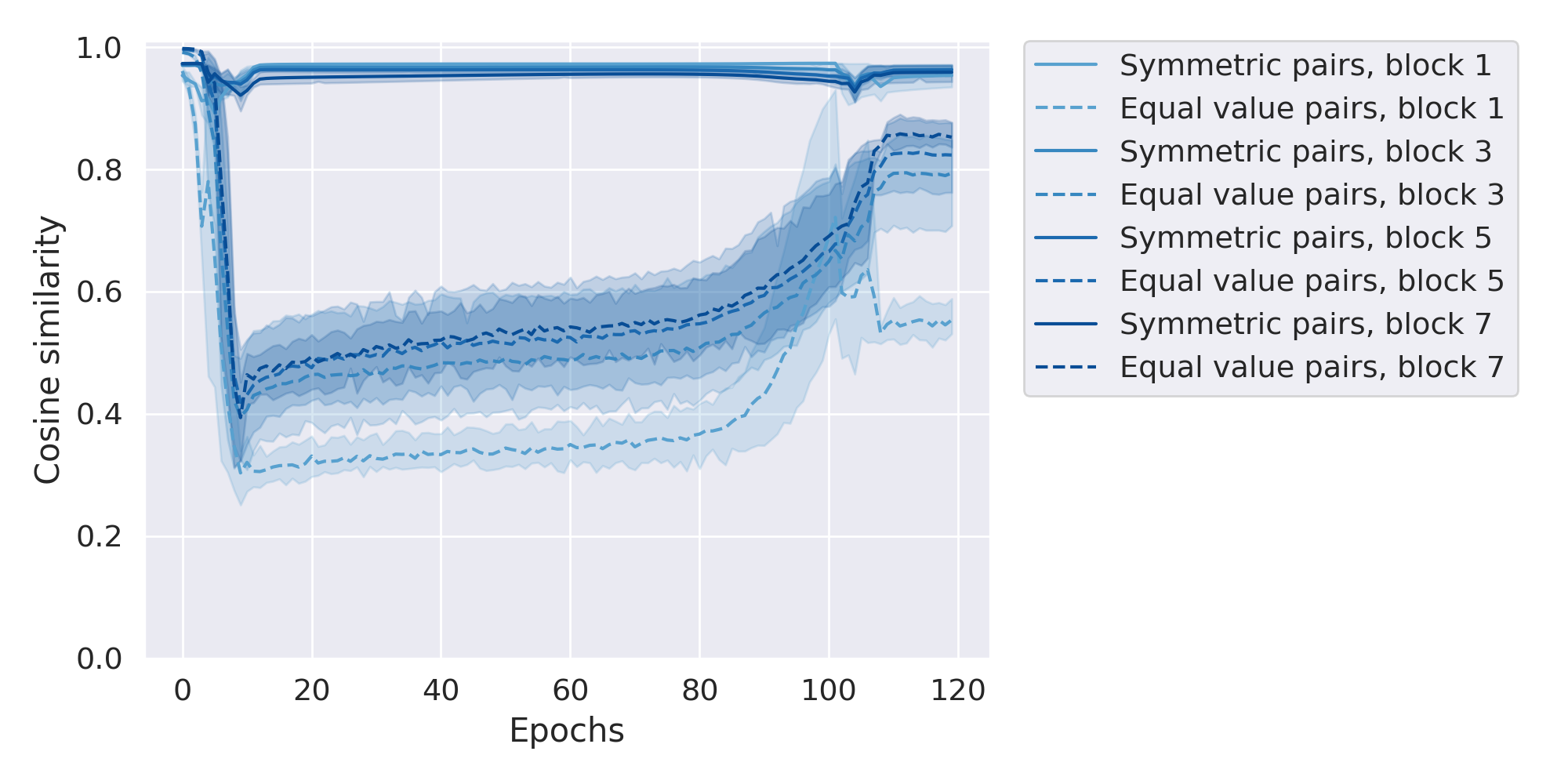}
    \caption{A plot of the symmetric representational similarity (solid lines) and equal value similarity (dashed lines) at different blocks of a decoder only transformer trained on the group $C_{100}$. Shading corresponds to 95\% confidence intervals over 5 random initializations.}
    \label{fig:cosine-similarity}
\end{figure}

Finally, examining performant model’s internal representations also reveal the fingerprints of commutativity with higher cosine similarity between pairs $g_1 \star g_2$ and $g_2 \star g_1$ relative to arbitrary equal value pairs. An illustration of this can be found in Figure \ref{fig:cosine-similarity}.

On their own, each of these tests provide relatively weak evidence that these models have internalized a notion of commutativity. Together however, they suggest that models capture a fundamental aspects of cyclic groups. 

\begin{mybluebox}
\textbf{Commutativity:} Transformers and MLPs appear to learn some notion of commutativity but this is not robust.
\end{mybluebox}

\subsection{The identity}
\label{subsect-identity}

Once one has identified the identity element $e$ in a group, computations involving $e$ are trivial for a human to perform. This is true even when one otherwise understands little else about the group. As such it is interesting to try to understand whether a neural network trained to predict the group operation also leverages the unique property that $g \star e = g$. We introduce two approaches to understand whether models learn the identity as a special and distinct element of the group like humans do. 

{\textbf{Identity accuracy:}} This simply involves tracking the accuracy on test examples of the form $g \star e$ or $e \star g$ in the test set. We then compare this to the global test accuracy. Substantial increases in identity accuracy relative to global accuracy may indicate a point in training where the model learns the identity.

As with symmetric consistency, we can also probe the extent to which a model has a strong concept of the identity element by testing whether it can generalize the properties of the identity to out of distribution examples. To do this we hold out a subset of elements $g_1',g_2',\dots,g_t' \in G$ so that none of these elements appear in the training set. The {\emph{OOD identity accuracy}} is then the accuracy of the subset of the test set of the form $g_j' \star e$ or $e \star g_j'$ for $1 \leq j \leq t$.

\textbf{Do models have a notion of the identity element?:}
In all our experiments, identity accuracy tracked the overall accuracy of the model closely giving no hint of a point where the model learned a specific prediction rule around the identity element. This is supported by results on OOD identity accuracy where no models were able to reliably predict that $g \star e = g$ or $e \star g = g$ if they had not seen $g$ during training. 

\begin{mybluebox}
\textbf{The identity element:} We were unable to surface any evidence that our neural networks recognize the identity as a special element of the group using the proposed techniques above. 
\end{mybluebox}

\subsection{Subgroup structure}
\label{sec:subgroup}
While understanding the identity element and commutativity offer obvious potential benefits for more efficient computation, the benefits of being able to distinguish different subgroups seems less clear. Nevertheless, analysis in works like \cite{mccracken2025uncovering} suggest that in some cases, the structure of certain subgroups may play a role in computation. For example, if one uses the Chinese remainder theorem for computation, one will naturally expose some subgroups of cyclic groups. Motivated by this, we propose the following tests to explore whether models learn to identify the subgroups of a group.

\textbf{Subgroup accuracy:} Analogous to identity accuracy, we can also look at {\emph{subgroup accuracy}}, the accuracy on pairs of elements $(h_1,h_2)$ belonging to the subgroup $H$. If we see significant changes in subgroup accuracy relative to global accuracy, this may correspond to a point in training where $f$ `learned' $H$ as a distinct part of $G$.

\textbf{Linear probing for subgroup membership:} We can test whether $f$ captures a distinct representation of $H$ by probing for subgroup membership on the hidden activations of $f$. More specifically, we can collect representation $v_{g_1,g_2}^k \in \mathbb{R}^{d_k}$ corresponding to $g_1 \star g_2$ at layer $k$, label them by whether $g_1,g_2 \in H$, and then train a linear probe to predict these labels.  

Note that in this test we should expect different behavior based on the data representation of transformers and MLPs. In the MLP case where input is a one-hot encoding of $g_1$ stacked on a one-hot encoding of $g_2$, this task should be easy in input space since the probe just needs to learn the $|H|$ indices corresponding to elements of $H$ in the first $|G|$ dimensions, the $|H|$ indices corresponding to elements of $H$ in the second $|G|$ dimensions, and be able to perform an `and' operation over these. We expect this task to become more challenging as $f$ transforms this initial representation of $(g_1,g_2)$ when computing $g_1 \star g_2$. In the case of transformers where the input is three tokens long: one token for $g_1$, one token for $g_2$, and one token for `$=$' and we predict $g_1 \star g_2$ from the third token, the task is impossible in input space (since the third token is the same for all input). It only becomes tractable as information from the first and second tokens representing $g_1$ and $g_2$ are transferred onto the third token via successive self-attention layers. In this situation if the representation of `$=$' retains information identifying the first argument as $g_1$ and the second argument as $g_2$, then it is possible that the subgroup could be learned via the same procedure described for the MLPs.

To better calibrate probe accuracy, we also train probes on pairs of elements from $R \subset G$, a random selection of $|H|$ elements of the group which do not form a subgroup. If linear probes can more effectively distinguish elements of $H$ than elements of $R$, we will have evidence that subgroup structure is captured in model hidden representations.

\textbf{Do models see subgroups?:} Unlike the process whereby a human might learn a group, first understanding a simpler subgroup and then building toward understanding the whole group, we find that across groups, subgroups, architectures, and hyperparameters, subgroup accuracy tracks global accuracy. This aligns with the trends for identity accuracy seen in Section \ref{subsect-identity}. 

\begin{figure}
    \centering
        \includegraphics[width = .49\textwidth]{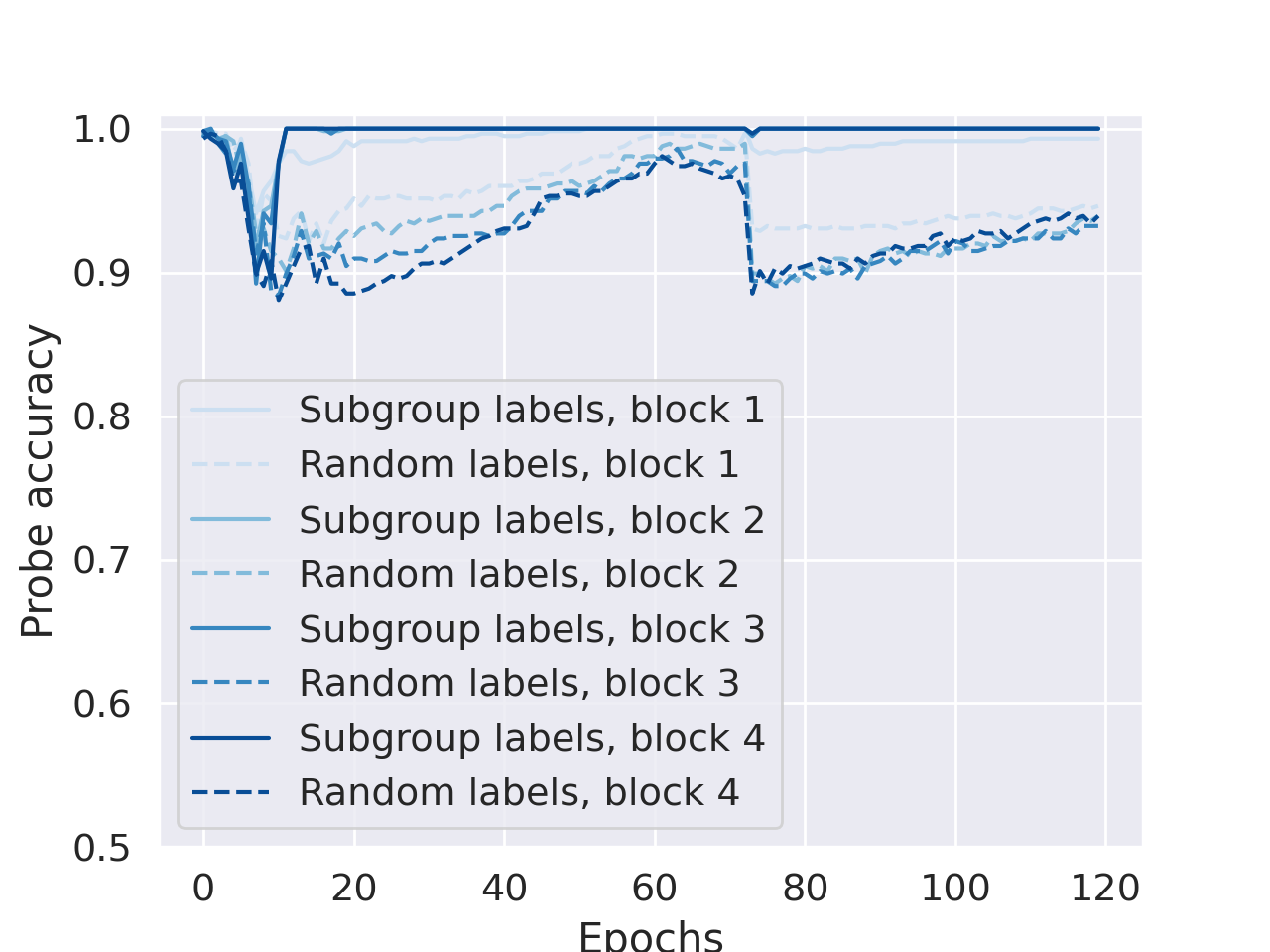}
        \includegraphics[width = .49\textwidth]{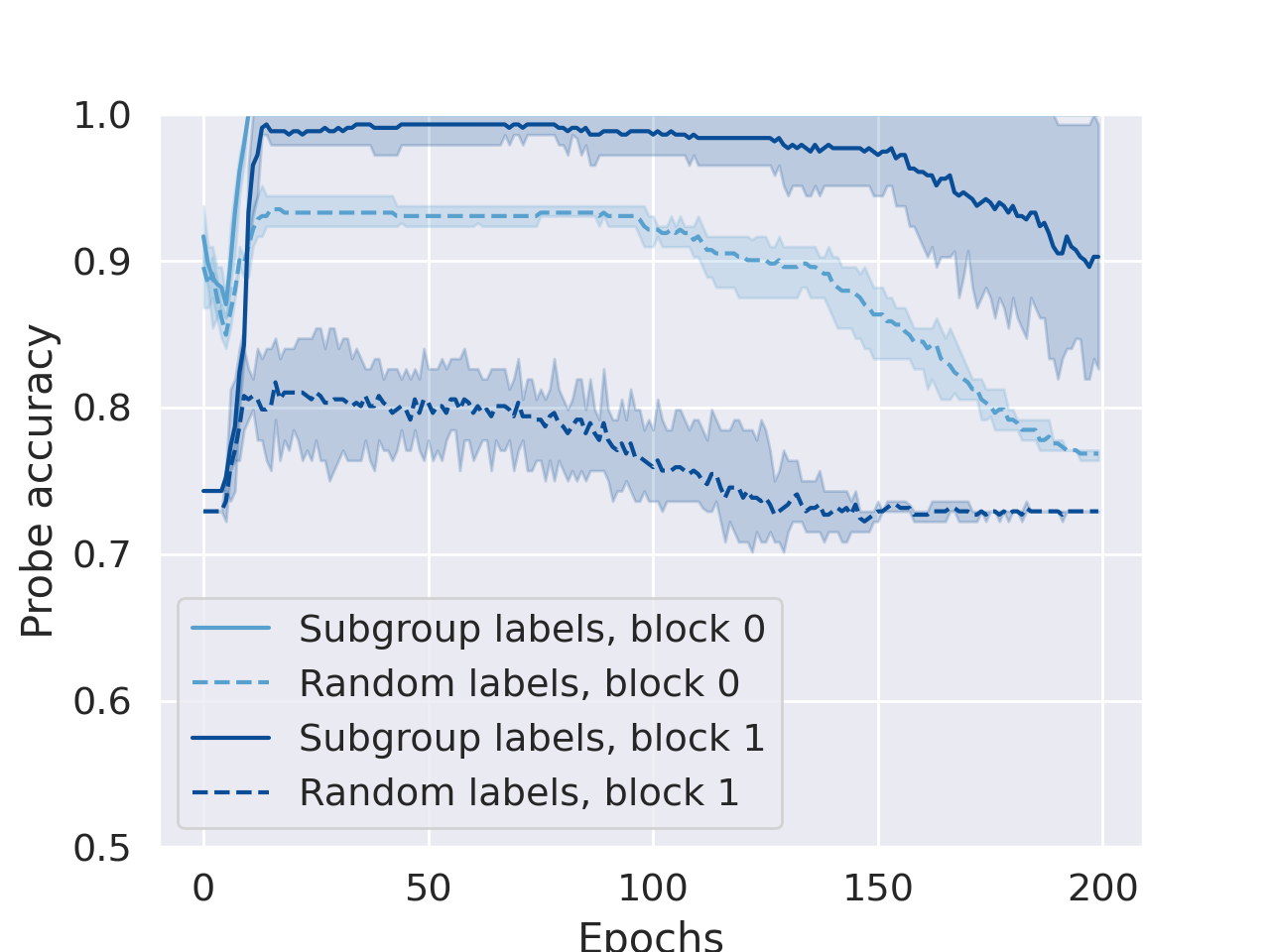}
    \caption{Linear probe performance on a \textbf{(Left)} transformer trained to predict the binary of $S_5$ and a \textbf{(Right)} MLP trained to predict the binary operation on $D_{30}$. Solid lines are probe accuracy when labels correspond to the alternating subgroup and rotation subgroup respectively, while the dashed line corresponds to a probe trained on a random labeling of the group. In the right plot, shading indicates 95\% confidence intervals over three random initializations (transformer performance on $S_5$ varied too much between random initializations to be useful in the case of the transformer). In the case of the MLP, probing was performed after each ReLU layer, in the transformer it was performed after each attention layer.}
    \label{fig:subgroup-probing}
\end{figure}

On the other hand, we find substantial evidence that performant models sometimes capture subgroup structure within their internal representations which we can access via the linear probing. We probe for several large subgroups of cyclic groups including the order $50$ and order $20$ subgroups generated in $C_{100}$, the subgroups generated by all rotations in $D_{30}$ and $D_{50}$ (order $30$ and $50$ respectively), and the order $60$ alternating subgroup in $S_{5}$. The accuracy of these probes, trained on both the true subgroup labels (solid lines) and random labels (dashed lines) over the course of training ($x$-axis) and at different layers (colors) are shown in Figure \ref{fig:subgroup-probing} for a transformer trained on $S_5$ (left), and an MLP trained on $D_{30}$ (right). Generally, probe performance on subgroups is substantially higher than performance on random subsets, especially at later layers where the impact of clean input encoding has diminished.

We stress that probe performance is variable across runs. Sometimes models achieve high accuracy predicting the group operation and yet their probe performance is close to guessing. 

\begin{mybluebox}
\textbf{Subgroup structure:} Subgroups can be more effectively detected via linear probing when compared to random subsets of the group that do not have algebraic structure.
\end{mybluebox}

\section{Discussion}
\label{sec:discussion}


\textbf{What mathematical structures tend to be captured by a small neural network?} This is an important question to answer as it helps us identify whether the small model paradigm will be useful for a particular AI for math research program. Based on the successes and failures described above, we suggest three intuitive rules of thumb to predict whether a given mathematical property $X$ is likely to be learned by a small model trained on a task $Y$. 
\begin{itemize}[leftmargin=.5cm]
\item Encoding $X$ allows the neural network to find a simpler solution applicable across many instances for the task $Y$. As an example, encoding commutativity and the trivial nature of the identity element can both lead to simpler solutions. But whereas commutativity applies to all instances, the special property of the identity only applies to $2|G|-1$ instances out of $|G|^2$.
\item $X$ fits into an existing interpretability framework. We found it hard to investigate whether models recognize the significance of the identity element because this property is challenging to formulate within interpretability and analysis techniques applicable to small models. For example, we could not devise a way of translating this concept into structure that we could look for in activation space.
\end{itemize}

\begin{table}
  \caption{Summary of algebraic structure captured in small LLMs and MLPs trained to perform a groups binary operation.}
  \label{sample-table}
  \centering
  \begin{tabular}{lll}
    \toprule
    Name & Type & Captured structure \\
    \midrule
    \multicolumn{1}{l}{Commutativity}                   \\
    \hspace{10pt}\small{Symmetric consistency} & \small{Learning dynamics} & \small{\color{ForestGreen}{Yes}}    \\
    \hspace{10pt}\small{OOD symmetric consistency}     & \small{Generalization}  & \small{\color{ForestGreen}{Yes}}  \\
    \hspace{10pt}\small{Symmetric representational similarity} & \small{Representation structure}  & \small{\color{ForestGreen}{Yes}}  \\
    \multicolumn{1}{l}{Identity}\\
    \hspace{10pt}\small{Identity accuracy} & \small{Learning dynamics}   & \small{\color{red}{No}}   \\
    \hspace{10pt}\small{Identity generalization} & \small{Generalization}    & \small{\color{red}{No}}  \\
    \multicolumn{1}{l}{Subgroup structure}\\
    \hspace{10pt}\small{Subgroup accuracy} & \small{Learning dynamics}    & \small{\color{red}{No}}  \\
    \hspace{10pt}\small{Hidden activation probing} & \small{Representation structure}    & \small{\color{ForestGreen}{Yes}} \\
    \bottomrule
  \end{tabular}
\end{table}

\textbf{Mathematical world models are sensitive to hyperparameter choice and initialization.} This may in part be due to the `fragile' nature of small algorithmic problems. Like others, we have found that the performance in these settings tends to be more sensitive to initial conditions and sources of randomness in training than other small-scale tasks (e.g., CIFAR10, MNIST). But even among training runs with identical hyperparameters (that converged) we found significant differences in the mathematical structures we were able to extract. For example, one trend that we noticed when training transformers on $S_5$ is that when the model converged earlier (after $~20-50$ epochs) the alternating subgroup was undetectable via linear probing but that when the model converged after training for longer ($>100$ epochs), the alternating subgroup was detectable. We hypothesize that cleaner solutions that tend to arise via grokking may capture more algebraic structure along the way. 

We suggest that the tests we offer here may be an interesting window into the fundamentally different solutions learned by a model over different training hyperparameters. Probing for mathematical world models may be a relatively cheap method of learning about the solution a model has converged to without reverse-engineering it (a labor-intensive process). Overall, we strongly recommend that when training models that will be used in downstream mathematical exploration, many different hyperparameters and initializations be explored, even after some models have been effectively trained.

\section{Related work}

This work presents a framework to understand whether narrow models trained to perform a single algebraic task—predicting a group operation—learn group-theoretic notions. This adds to the growing literature on world models of neural networks, which has focused largely on sequence models in strategy game environments, and more recently LLMs \cite{Mitchell2023AIsCO, Li2022EmergentWR, Nanda2023EmergentLR, rohekar2025a}, while the algebraic world models of narrow models trained on mathematics tasks are much less explored. 

Closest to our setting are mechanistic studies that reverse-engineer how small MLP and transformer models learn group operations.  \cite{nanda2023progressmeasuresgrokkingmechanistic, Zhong2023TheCA, chughtai2023toy, stander2023grokking, wu2024unifying, mccracken2025uncovering}. These works have often found that the learned algorithms use group-theoretic structure, although they sometimes differ in the specific algorithms they reverse-engineer \cite{chughtai2023toy, stander2023grokking}. Our work is complementary, asking whether we can extract evidence of abstract notions like commutativity or subgroup structure, regardless of whether we can extract the precise algorithm used.  

Our work is motivated by the potential to use AI systems for mathematical discovery. This is a rapidly growing field, including using AI to generate proofs \cite{Yang2024FormalMR}, discover counterexamples \cite{Wagner2021ConstructionsIC}, and find mathematical constructions \cite{Charton2024PatternBoostCI, Alfarano2024GlobalLF, Yip2025TransformingCC}. Some of this work trains a model to solve a task closely related to the problem, and then relies on expert probing to extract mathematical insight \cite{Davies2021AdvancingMB, hemachines}. In contrast, we take a step back to ask whether narrow models learn abstractions needed in order to gain useful insights.

\section{Limitations}
\label{sect:limitations}
This work uses the elementary setting of finite groups to explore the prospects for using machine learning models to surface interesting mathematics that falls outside the task a model was designed to solve. Given the set-up, we know in advance the kinds of structures we are looking for. As such, we sidestep the main technical challenge in this research program (knowing what to look for in the first place). However, we hope that our results which show that interesting structure does appear in models, will support the idea that this is a worthwhile research direction that remains accessible to researchers with medium to small computational resources.

\section{Conclusion}

In this paper we explore the question of whether neural networks trained on a simple mathematical task, predicting the binary operation that defines a group, capture interesting structure not specified in the task itself. Our results provide support for the idea that even small neural networks can learn interesting structure that goes beyond simply solving the task. We hope that this will motivate the use of more advanced analytical techniques in the field of AI for math. We see trained models as vehicles for mathematical exploration, particularly in settings where the scale of the computation is potentially beyond what a human can easily perform.

Despite our positive results, we see the most significant technical challenge in the widespread use of this paradigm as being able to effectively extract insights from a model when one does not already know what they are. This is, and should continue to be, a major research effort in the coming years.

\bibliographystyle{plainnat}
\bibliography{references}


\appendix

\section{Hyperparameters} \label{sect:hyperparameters}
We provide the hyperparameters used to generate the plots below.
\begin{itemize}
    \item In order to generate the plots in Figures \ref{fig:commutativity} and \ref{fig:cosine-similarity} we used decoder-only transformers with: 4 attention blocks and 4 MLP blocks, residual stream dimension $1,000$, $8$ attention heads, learning rate $0.0001$, weight decay $0.001$. We used $80\%$ of all $10,000$ instances for training and $20\%$ for test.
    \item For Figure \ref{fig:subgroup-probing} (left), we used 3 decoder-only transformers with: 4 attention blocks and 4 MLP blocks, residual stream dimension $1,000$, $8$ attention heads, learning rate $0.0001$, weight decay $0.001$. We used $80\%$ of all $14,400$ instances for training and $20\%$ for test.
    \item For Figure \ref{fig:subgroup-probing} (right), we used 3 MLPs of depth 2 and width $1,000$, learning rate $0.001$, and weight decay $0.0005$. We used $80\%$ of all $3,600$ instances for training and $20\%$ for test.    
\end{itemize}

\end{document}